\documentclass[10pt,twocolumn,letterpaper]{article}

\usepackage[algorithms]{wacv}      

\definecolor{wacvblue}{rgb}{0.21,0.49,0.74}
\usepackage[pagebackref,breaklinks,colorlinks,allcolors=wacvblue]{hyperref}
\usepackage{algorithm}
\usepackage{algpseudocode}
\usepackage{multirow} 

\def\wacvPaperID{2279} 
\def\confName{WACV}
\def\confYear{2027}

\title{When Depth Is Better Told Than Shown: Depth-Ordinal Prompting for Vision-Language Spatial Reasoning}

\author{
  Quynh Vo \quad Phuc Dao \quad Cong-Duy Nguyen \quad Thong Nguyen\thanks{Corresponding author} 
  \\
  National University of Singapore \quad Center of AI Research, VinUniversity \\ \href{mailto:thong.nguyen@u.nus.edu}{thong.nguyen@u.nus.edu}}

\begin{document}
\maketitle

\begin{abstract}
Vision-language models (VLMs) are expected to reason about physical space---which object is closer, what lies behind what, and how objects are arranged in 3D---yet they still struggle with such spatial judgments. A natural remedy is to show the model a depth map, but we find that this can make performance worse. We show that depth is not absent: it reaches the language model, but becomes difficult to access for downstream reasoning, while rendered pseudo-depth maps act as noisy auxiliary images that frozen VLMs cannot easily regulate. We propose Depth-Ordinal Prompting (DOP), a training-free method that converts monocular depth into a single question-targeted ordinal text cue at the queried objects, without adding a depth image, training a module, injecting features, or using labels. Our key finding is form dependence: the same depth signal can hurt when shown as an image but help when told as text.Across benchmarks, models, and depth estimators, DOP improves spatial reasoning when pseudo-depth provides reliable object-level ordering and remains largely neutral in strong original-image regimes. It is also competitive with the strongest training-free depth-prompting alternative while being simpler and more targeted. 

\end{abstract}

\section{Introduction}
\label{sec:intro}

\begin{figure}[t]
    \centering
    \includegraphics[width=\linewidth]{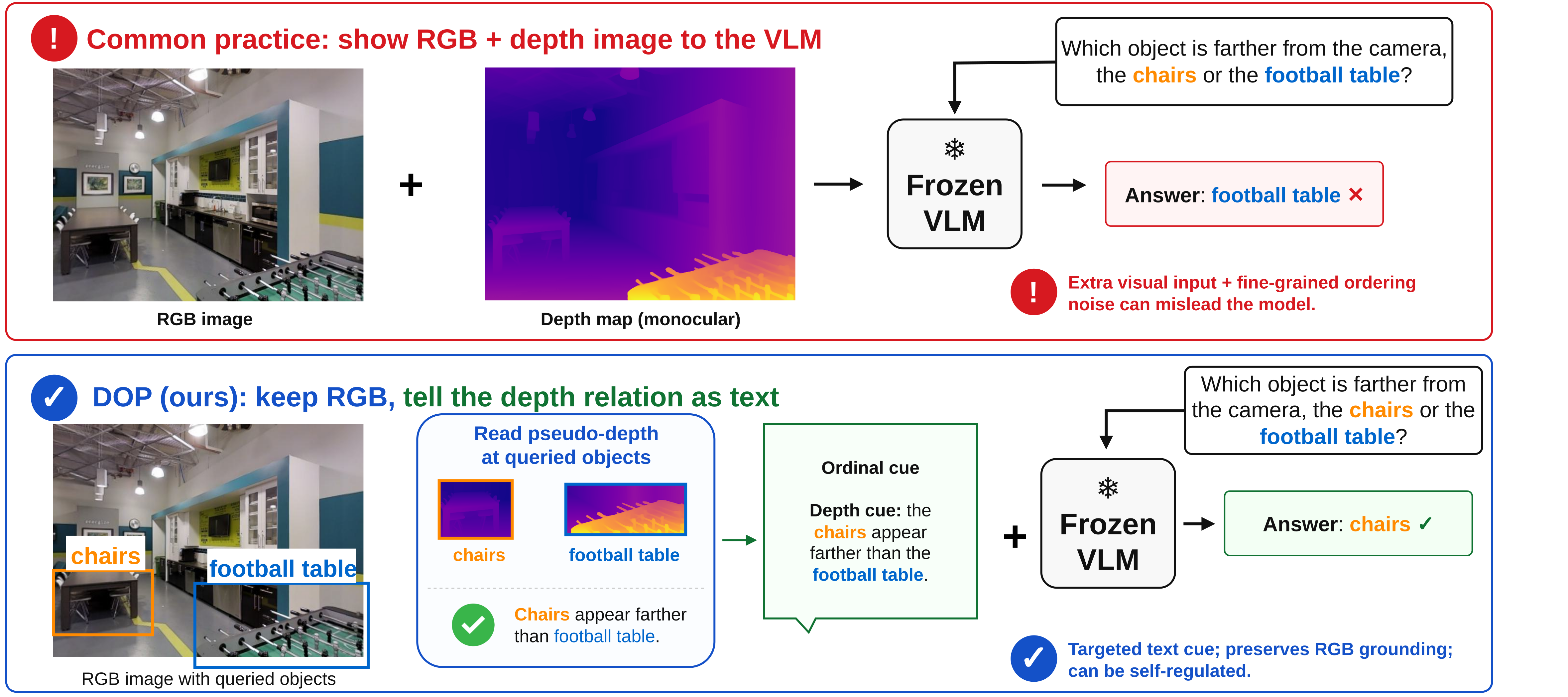}
    \caption{\textbf{Tell, don't show.} The same monocular depth signal can hurt when shown as an additional image but help when converted into a targeted ordinal text cue. DOP keeps the original color image and communicates only the queried-object depth relation, allowing the frozen VLM to weigh the cue against its original-image evidence.}
    \label{fig:teaser}
\end{figure}

Vision-language models (VLMs) are increasingly expected to reason about physical space: which object is closer, what lies behind what, and how objects are arranged in 3D. Such capabilities are central to robotics, embodied AI, and assistive perception~\cite{spatialvlm,spatialbot,spatialrgpt}. Yet spatial reasoning remains a persistent weakness. Recent benchmarks show that even strong VLMs struggle with relative depth, direction, and object-level spatial relations~\cite{blink,cvbench,vsr,embspatial,threedsrbench}. A natural remedy is to provide geometry explicitly, for example by appending a monocular depth map to the original color image or by training models with depth- and 3D-aware supervision~\cite{spatialvlm,spatialbot,spatialrgpt,roborefer,sdvlm,ssr}. Surprisingly, we find that the simplest version of this remedy often makes performance \emph{worse}: when pseudo-depth is rendered as an additional image, frozen VLMs can degrade substantially on point- and object-level spatial reasoning. Fig.~\ref{fig:teaser} previews this \emph{form contrast}: rendering depth as another image can mislead the frozen VLM, whereas expressing the queried-object relation as a short text cue can make the same geometric information easier to use. This raises a basic question: if depth is useful, why does showing depth hurt?

This contrast suggests that the issue is not simply the absence of depth, but the \emph{interface} through which depth is made usable by a frozen VLM. We therefore ask whether the model fails because it never encodes depth, or because depth becomes difficult to use later in the computation. To answer this, we conduct layer-wise probing of the model's hidden states, as summarized in Fig.~\ref{fig:layersweep}. We find that early visual features contain a clear depth signal, and that this signal reaches the first language-model layers with little loss. However, as computation proceeds, the same signal becomes hard to read with a simple linear probe, even though a non-linear probe can still recover it. In other words, depth is not erased; it becomes entangled in a form that is less accessible to the model's downstream reasoning. A causal intervention further supports this view: when depth is re-supplied in a linearly readable form, accuracy on indoor depth ordering improves substantially and approaches a sensor-depth reference. Thus, the bottleneck is not a complete lack of geometry, but a mismatch between how depth is encoded and how the frozen model can use it.

We then diagnose why depth \emph{maps} are a poor interface. Depth helps only when reliable at the task's granularity: GT sensor depth helps region-level reasoning, while monocular pseudo-depth hurts fine-grained object-level comparisons---and not because of raw accuracy alone, since region cues tolerate large noise (averaging cancels errors) while object-level answers flip under small ordering mistakes. A rendered map also becomes an extra visual input the frozen VLM must align, de-colormap, and selectively ignore; a marker control rules out reference-marker loss. The useful signal is thus not the full map or raw metric depth, but the \emph{decision-relevant ordinal relation} at the queried objects.

Guided by this diagnosis, we propose Depth-Ordinal Prompting (DOP), a simple training-free interface for communicating depth to frozen VLMs. Given an original color image, a spatial question, and the queried object regions, DOP reads monocular pseudo-depth at the object boxes using an off-the-shelf depth estimator~\cite{depthanythingv2,dpt}, converts the result into the question-relevant ordinal relation, and appends it to the original prompt as a single text cue, e.g., ``the \textit{chair} appears closer than the \textit{table}.'' DOP does not add a depth image, inject features, train an auxiliary module, or require labels; it keeps the original image and communicates only the depth relation needed for the decision. This yields our central finding of \textbf{form dependence}: the same depth signal can hurt when shown as an image, but remain useful when told as text. Wrong-cue tests and attention-knockout interventions show the cue is empirically self-regulating: the model relies on it under visual uncertainty and discounts it when the original-image answer is already confident---so an always-on cue helps where visual reasoning is weak yet stays neutral when it is strong, making an explicit gate unnecessary in our setting.


\noindent Our contributions are threefold:
\begin{itemize}
    \item We diagnose depth use in frozen VLMs, showing that depth is not absent but becomes difficult to access in later language-model layers. We further identify why pseudo-depth images can hurt: they introduce fine-grained ordering errors, cross-image alignment, colormap interpretation, and visual-token burden.
    \item We introduce \textbf{Depth-Ordinal Prompting (DOP)}, a training-free method that communicates depth as a single question-targeted ordinal text cue at the queried objects, without adding a depth image, training a module, injecting features, or using labels.
    \item Across benchmarks, models, depth estimators, and ablations, we show that DOP improves spatial reasoning when pseudo-depth provides reliable object-level ordering and remains largely neutral when original-image reasoning is already strong.
\end{itemize}

\section{Related Works}
\label{sec:related}

\subsection{Spatial and Geometric Reasoning in VLMs}

Recent benchmarks show that VLMs remain weak at spatial reasoning, including relative depth, direction, metric distance, and object-level spatial relations~\cite{blink,cvbench,vsr,embspatial,threedsrbench,predict_trust}. These failures have motivated depth- and 3D-augmented VLMs. SpatialVLM~\cite{spatialvlm} uses synthetic 3D QA data; SpatialBot~\cite{spatialbot} trains on paired RGB-depth supervision; SpatialRGPT~\cite{spatialrgpt} and RoboRefer~\cite{roborefer} inject depth or 3D features through trained modules; SD-VLM~\cite{sdvlm} adds depth positional encoding; and SSR~\cite{ssr} distills depth-guided rationales into a trained model. These methods demonstrate that geometry is useful, but they require model-specific training, data curation, or feature-level integration. In contrast, DOP targets the harder plug-and-play setting: the VLM is frozen, no module is trained, and depth is communicated only through the standard text prompt.

Closer to our setting are training-free spatial and depth prompting methods. ByDeWay~\cite{bydeway} converts monocular depth into global layered scene captions for a frozen VLM; SpatialPIN~\cite{spatialpin} prompts with reconstructed 3D priors; SpatialPrompt~\cite{spatialprompt} uses reference objects to elicit quantitative spatial reasoning; and visual-prompting methods such as Set-of-Mark~\cite{som}, SCAFFOLD~\cite{scaffold}, and Coarse Correspondences~\cite{coarse_corr} modify the input image with marks or correspondences. DOP differs by using only a single question-targeted ordinal relation at the queried objects. It adds no pixels, performs no 3D reconstruction, and avoids global scene captions, making it a minimal interface for testing how frozen VLMs use auxiliary depth.
While prior work establishes that geometry can benefit VLMs, we show that its \emph{interface} is equally critical. The same monocular depth signal can become harmful when rendered as an additional image, yet become useful when distilled into a question-targeted ordinal text cue.

\subsection{Interpreting and Regulating Auxiliary Evidence}

Our analysis connects to mechanistic interpretability work that probes or intervenes on model representations. Prior studies decompose vision-language representations~\cite{clip_decomposition}, probe attention heads for hallucination~\cite{vibprobe}, and edit activations at inference time~\cite{iti}. We use layer-wise probing and causal feature injection in a similar diagnostic spirit, but focus specifically on depth. These analyses show that depth is not absent from the model; rather, it becomes difficult to access in later language-model layers. This motivates DOP as the deployed method, while the feature injection serves only as evidence for where the bottleneck lies.

DOP also relates to reliability-gated fusion and selective prediction. Prior work uses confidence to gate multimodal fusion~\cite{multimodal_dynamics}, detect uncertainty or hallucination~\cite{vl_uncertainty,recoverr}, or decide when to retrieve external evidence~\cite{adaptive_retrieval}. Calibration and knowledge-conflict studies further examine whether models can report confidence or follow conflicting context~\cite{justask_calibration,knowledge_conflict}. These works motivate explicit control over when auxiliary evidence should be used. Our finding is different: when depth is communicated as a targeted text cue, an explicit gate is not necessary in our setting. The frozen VLM empirically tends to rely on the cue when original-image reasoning is uncertain and discount it when the original-image answer is already confident.
Rather than proposing another confidence gate or decoding rule, we identify the \emph{interface} of auxiliary evidence as a key factor in whether a frozen VLM can regulate it. In our setting, depth is difficult to discount when shown as an additional image, but becomes more controllable when expressed as a targeted text cue.

\section{Diagnosing the Depth Interface Failure}
\label{sec:analysis}

\begin{figure}[t]
\centering
\includegraphics[width=\linewidth]{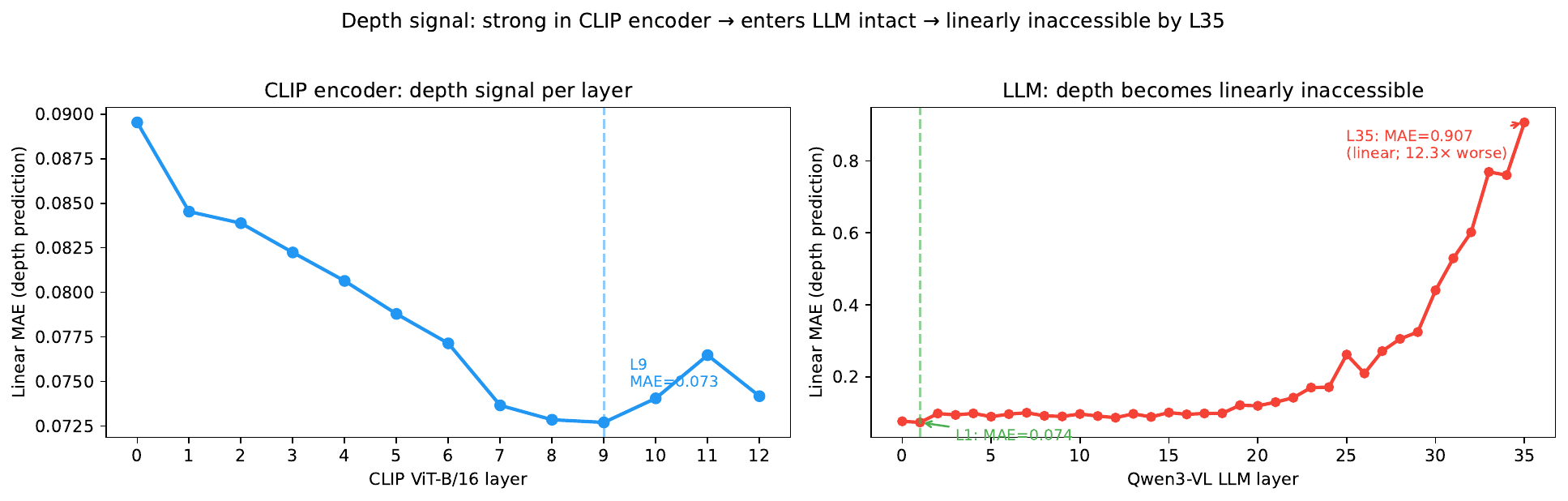}
\caption{\textbf{Layer-wise depth decodability.} A linear probe's depth-prediction error grows $12.3\times$ across Qwen3-VL's LM layers (L1 MAE $0.074\to$ L35 $0.907$), even though a standalone CLIP ViT-B/16 encoder carries a strong causal depth signal (L9 MAE $0.073$) also present at the first LM layer ($0.074$). A non-linear probe recovers depth at L35 (MAE $0.074$), so depth is not destroyed but becomes linearly inaccessible.}
\label{fig:layersweep}
\end{figure}

\subsection{Depth persists but becomes linearly inaccessible}
\label{sec:probe}

Do spatial failures arise because the model never encodes depth, or because depth becomes hard to use downstream? We fit a per-patch linear regressor from hidden states to ground-truth depth (MAE on held-out patches). A vision encoder carries a strong depth signal: a standalone CLIP ViT-B/16 (the ESDI feature source) peaks at L9 (MAE $0.073$), and it is causal---ablating the depth direction raises MAE $83\%$ vs.\ ${<}1\%$ for a random one.

The signal enters Qwen3-VL's language model nearly intact (first-layer MAE $0.074$), but linear decodability then degrades sharply, reaching MAE $0.907$ at the final layer---a $12.3\times$ increase (Fig.~\ref{fig:layersweep}). This is not destruction: a non-linear MLP probe still recovers depth to near encoder-level accuracy at the final layer (MAE $0.074$). Depth thus persists in the residual stream but becomes nonlinearly entangled---inaccessible to a simple linear readout.


To test whether this linear inaccessibility is \emph{causally} tied to spatial failure, we perform an embedding-space depth injection (ESDI): we write clean, linearly-structured depth features from a vision encoder directly onto the language model's first-layer visual-token positions, bypassing the rendered-image interface. Were the failure a perception gap, re-supplying depth this way should not help; instead, on the NYU test set ESDI improves indoor depth ordering by $+13.1$pp over the original-image baseline (Table~\ref{tab:esdi}). The gain is sharpest on \emph{conflict} pairs---items where a vertical-position prior (objects lower in the image are assumed closer) contradicts the true depth order---where ESDI matches the sensor-depth condition within $0.2$pp. The language model can therefore use depth when it is reintroduced in an accessible form. ESDI is a diagnostic, \emph{not} a deployable baseline competing with DOP: the injected features are CLIP-encoded from the \emph{ground-truth} depth map and mapped into the LM's token space by a \emph{fitted} linear projection, so ESDI presupposes GT-quality depth and feature-level access. It only tests whether the LM \emph{can} use depth when it is supplied in a linearly accessible form, and thereby motivates DOP: communicate only the decision-relevant depth relation, in a form the frozen model can read.

\begin{table}[t]
\centering
\caption{ESDI causal probe (NYU, GT depth; full 654-image test set, $n{=}5669$ pairs).}
\label{tab:esdi}
\resizebox{\linewidth}{!}{%
\begin{tabular}{@{}lll@{}}
\toprule
Condition & Acc. (\%) & $\Delta$ / note \\
\midrule
Original image only & $75.5$ & --- \\
GT depth (image) & $89.8$ & $+14.2$ \\
\textbf{ESDI injection} & $\mathbf{88.6}$ & $\mathbf{+13.1}$ \\
ESDI (MLP proj.) & $89.6$ & $+14.1$ \\
ESDI conflict pairs & $86.3$ & matches sensor-depth ceiling (conflict subset) \\
\bottomrule
\end{tabular}
}
\end{table}

\subsection{Why rendered depth maps hurt}
\label{sec:whymapshurt}

\begin{figure}[t]
\centering
\includegraphics[width=\linewidth]{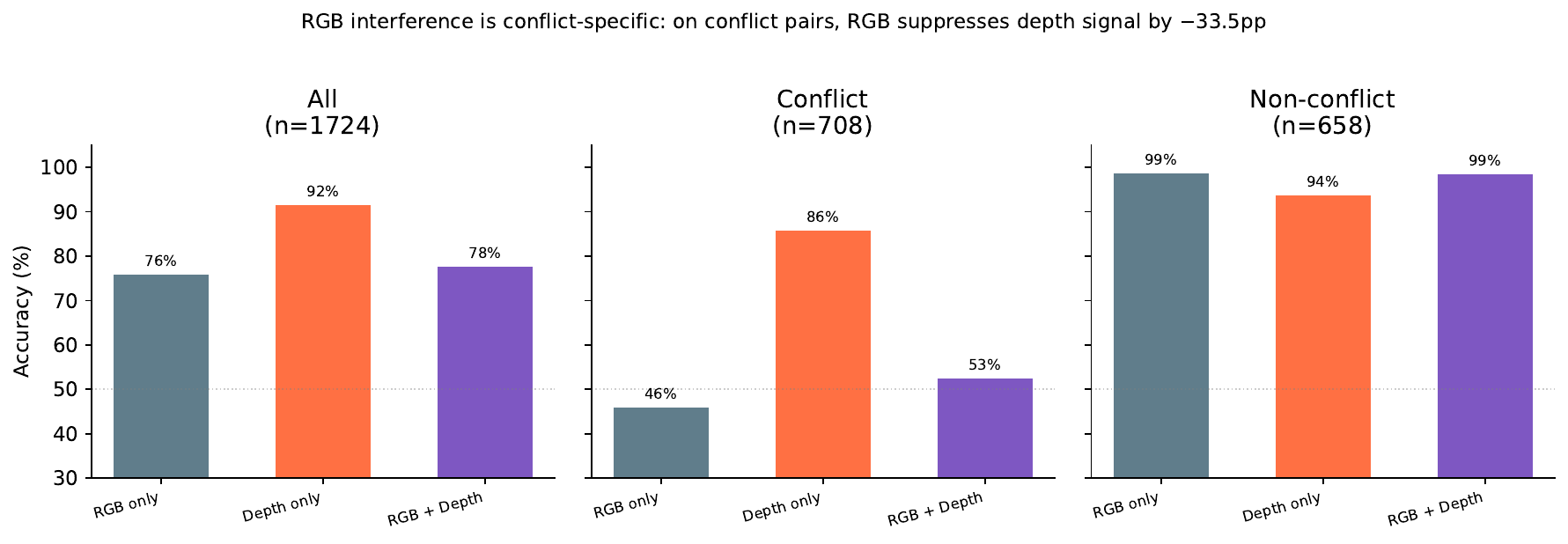}
\caption{\textbf{Interference is conflict-specific.} When a vertical-position prior agrees with sensor depth, adding the original image to a depth map is mostly harmless. When they conflict, performance drops by $33.5$pp, suggesting that the model shifts toward the visual prior.}
\label{fig:conflict}
\end{figure}

If the model can use depth, why does showing it a depth map often hurt? Table~\ref{tab:gradient} reports a quality gradient comparing the original image against the rendered depth map alone across seven spatial benchmarks; the complementary case of \emph{combining} the original image with depth is analyzed as an interference diagnostic in Fig.~\ref{fig:conflict}. The results reveal a sharp quality and granularity gradient. Ground-truth sensor depth helps on region-level tasks such as NYU Depth V2 and KITTI, while monocular pseudo-depth substantially hurts on fine-grained point- and object-level benchmarks such as BLINK, CV-Bench, VSR, and EmbSpatial.

\begin{table}[t]
\centering
\caption{Depth-map encoding: quality gradient ($\Delta$ vs.\ original-image only); the ``Depth img.''\ column is the depth map shown \emph{alone} (replacing RGB), distinct from the RGB+heatmap condition in the baseline tables. GT sensor depth helps region-level tasks, while monocular pseudo-depth hurts point/object-level tasks.}
\label{tab:gradient}
\resizebox{\linewidth}{!}{%
\begin{tabular}{@{}llrrrr@{}}
\toprule
Benchmark & $n$ & Depth src.& Orig. image & Depth img. & $\Delta$ \\
\midrule
NYU Depth V2 & 5669 & GT Kinect & $75.5$ & $89.8$ & $\mathbf{+14.2}$ \\
KITTI dense & 9900 & GT LiDAR & $82.6$ & $86.9$ & $\mathbf{+4.3}$ \\
BLINK Rel-Depth & 124 & pseudo & $80.7$ & $53.2$ & $\mathbf{-27.4}$ \\
CV-Bench D+D & 1200 & pseudo & $92.8$ & $70.1$ & $\mathbf{-22.7}$ \\
VSR (depth) & 878 & pseudo & $83.1$ & $59.6$ & $\mathbf{-23.6}$ \\
EmbSpatial & 1206 & pseudo & $69.3$ & $52.9$ & $\mathbf{-16.4}$ \\
3DSR-Bench & 5072 & pseudo & $60.7$ & $54.2$ & $\mathbf{-6.4}$ \\
\bottomrule
\end{tabular}
}
\end{table}

\begin{figure}[t]
\centering
\includegraphics[width=\linewidth]{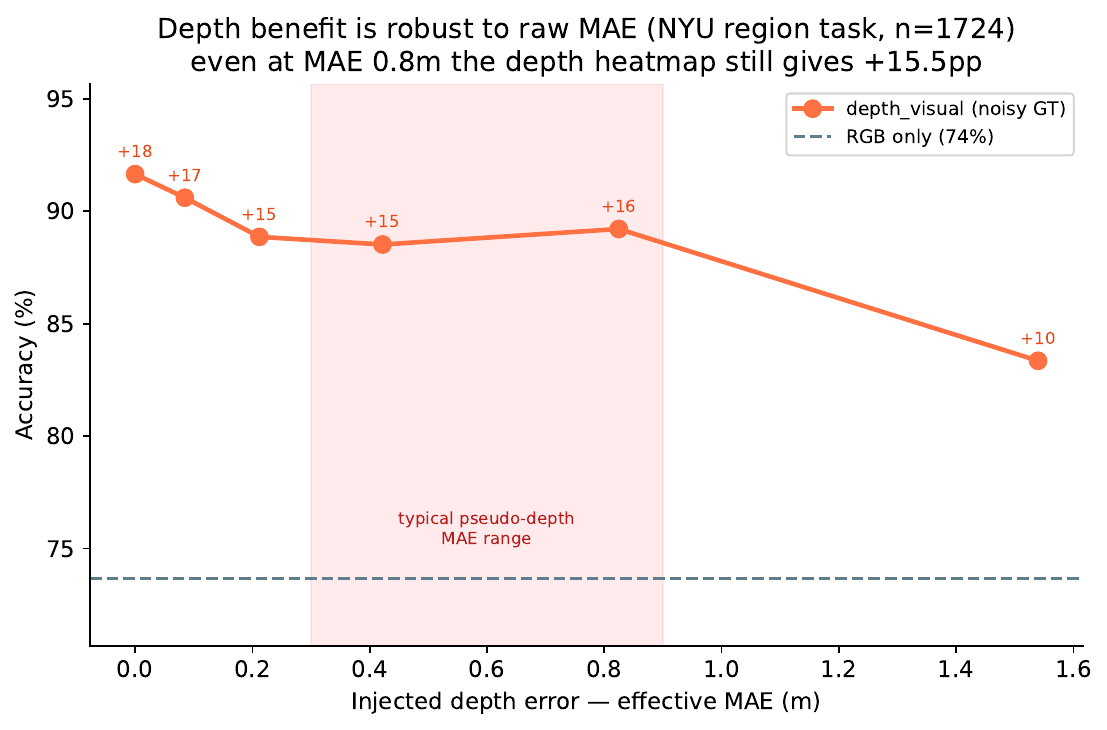}
\caption{\textbf{Depth-noise robustness.} Region-level depth benefits are robust to substantial zero-mean noise: even at an effective MAE of $0.8$\,m, corrupted GT depth still improves accuracy by $15.5$pp. This indicates that the pseudo-depth failures on object-level benchmarks are not explained by global MAE alone.}
\label{fig:noise}
\end{figure}

Two diagnostics clarify the source of this harm. First, interference is conflict-specific. When a vertical-position prior agrees with sensor depth, adding the original image to a depth map is mostly harmless. When the two conflict, performance drops by $33.5$pp, suggesting that the model shifts away from the depth map toward the visual prior (Fig.~\ref{fig:conflict}). Second---and key to resolving an apparent paradox---the harm comes from the error's \emph{structure} and the task's \emph{granularity}, not from global depth accuracy. The injected Gaussian corruption is \emph{zero-mean} and read at the \emph{region} level (NYU/KITTI), so region-mean pooling cancels it; the $+15.5$pp benefit therefore persists even at $0.8$\,m MAE (Fig.~\ref{fig:noise}). Monocular pseudo-depth is different on both axes: it makes \emph{structured, non-zero-mean} errors that pooling cannot average away, and the failing benchmarks are \emph{object}-level, hinging on the fine-grained ordering of small adjacent boxes where a single such error flips the answer. So robustness to zero-mean region-level noise and fragility to systematic object-level error are consistent, not contradictory.

Together, rendered depth maps fail for two reasons: pseudo-depth introduces fine-grained ordering errors at the queried objects, and the render itself is an extra visual input the frozen model cannot easily down-weight. We rule out reference-marker loss---drawing the queried boxes onto the depth map does not recover accuracy (App.~\textcolor{wacvblue}{G}). The fix is a different interface: communicate the decision-relevant ordinal relation at the queried objects \emph{as text}.

\section{Depth-Ordinal Prompting}
\label{sec:method}

\begin{figure*}[t]
\centering
\includegraphics[width=\textwidth]{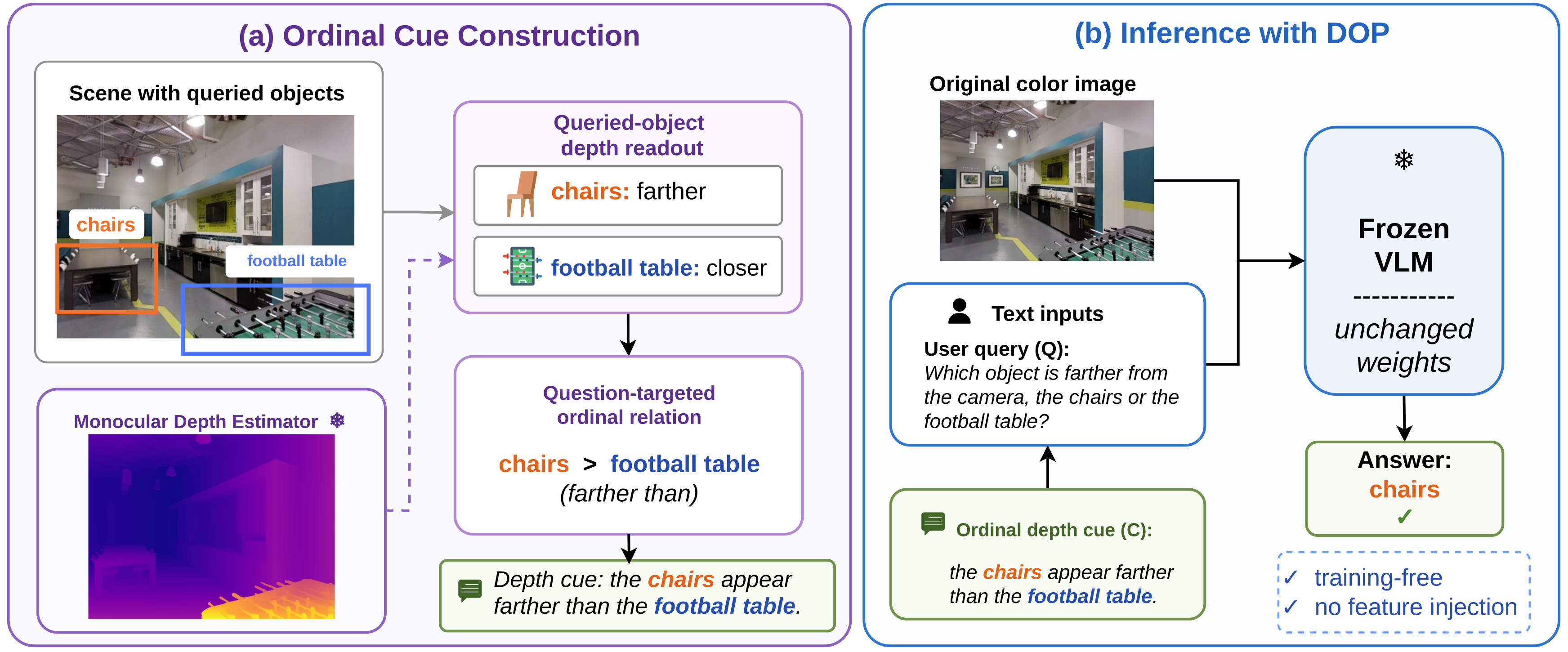}
\caption{\textbf{Overview of Depth-Ordinal Prompting (DOP).}
DOP treats depth as an internal signal rather than an additional visual input. Given the original image, the spatial question, and the queried object regions, DOP estimates monocular pseudo-depth, reads depth only at the queried objects, converts the readout into a question-targeted ordinal relation, and verbalizes it as a short text cue. The frozen VLM receives the original image, the question, and the ordinal cue; the depth map itself is never shown to the model.}
\label{fig:overall_method}
\end{figure*}

The analysis in Sec.~\ref{sec:analysis} suggests that rendering depth as an auxiliary image forces the VLM into complex visual alignment and colormap interpretation. As illustrated in Fig.~\ref{fig:overall_method}, \textbf{Depth-Ordinal Prompting} (DOP) bypasses this by compressing dense pseudo-depth into a single question-targeted ordinal text cue. DOP is training-free, label-free, and model-agnostic, requiring no visual feature injection or auxiliary module training.

\subsection{Formalization and Interface Bottleneck}
\label{sec:dop-formalization}
Let $I$ be the color image, $q$ a spatial question, and $\mathcal{O}=\{o_i\}_{i=1}^{k}$ the queried objects with bounding regions $\mathcal{B}=\{B_i\}_{i=1}^{k}$. A standard frozen VLM predicts an answer via $\hat{y}_{\mathrm{RGB}} = \arg\max_y p_{\theta}(y \mid I, q)$. DOP augments this text prompt while keeping the VLM unchanged. 

Instead of feeding a dense depth map $D = g(I)$ from an off-the-shelf estimator $g$ directly into the visual stream, DOP applies a question-conditioned information bottleneck:
\[
D \xrightarrow{\quad \mathcal{B},\,q \quad} \{d_i\}_{i=1}^{k} \xrightarrow{\quad \pi_q \quad} r_q \xrightarrow{\quad \tau \quad} c_q.
\]
First, dense pixel depth is aggregated into an object-level readout to mitigate local noise:
\[
d_i = \operatorname{Pool}\{D(u) \colon u \in B_i\},
\]
where $\operatorname{Pool}$ computes the mean depth within region $B_i$. We standardize all estimators so that larger values indicate closer distance (the a-priori DepthAnything-v2 convention).

Next, a question-dependent operator $\pi_q$ maps the object depths into a targeted ordinal relation $r_q = \pi_q(d_1,\dots,d_k)$. For a binary comparison between objects $o_i$ and $o_j$, the relation resolves to:
\[
r_{ij} = \begin{cases}
o_i \text{ is closer than } o_j, & d_i > d_j, \\
o_i \text{ is farther than } o_j, & d_i < d_j.
\end{cases}
\]
Finally, a verbalizer $\tau$ maps $r_q$ to a short natural-language cue $c_q = \tau(r_q)$ (e.g., ``the \textit{chair} appears farther than the \textit{table}''). The final DOP prediction is obtained via:
\[
\hat{y}_{\mathrm{DOP}} = \arg\max_y p_{\theta}(y \mid I, q \oplus c_q),
\]
where $\oplus$ denotes prompt concatenation. Crucially, the VLM only processes $(I, q \oplus c_q)$. By filtering out scene-wide irrelevant geometry and raw metrics, this linguistic bottleneck exposes only the exact core ordinal constraint needed for the decision.

\begin{algorithm}[t]
\caption{Depth-Ordinal Prompting (DOP)}
\label{alg:dop}
\begin{algorithmic}[1]
\Require Original image $I$; spatial question $q$; queried objects $\mathcal{O}=\{o_i\}_{i=1}^{k}$ with regions $\mathcal{B}=\{B_i\}_{i=1}^{k}$; frozen VLM $f_\theta$; monocular depth estimator $g$.
\State $D \gets g(I)$ \Comment{Pseudo-depth used internally only}
\State Convert $D$ to a common convention where larger values mean closer.
\For{each queried object $o_i$}
\State $d_i \gets \operatorname{Pool}\{D(u) \colon u \in B_i\}$ \Comment{Object-level depth readout}
\EndFor
\State $r_q \gets \pi_q(d_1,\dots,d_k)$ \Comment{Question-targeted ordinal relation}
\State $c_q \gets \tau(r_q)$ \Comment{Verbalize as a short text cue}
\State \Return $f_\theta(I, q \oplus c_q)$ \Comment{One VLM inference with cued prompt}
\end{algorithmic}
\end{algorithm}

\subsection{Design Rationale}
\label{sec:doprationale}
DOP is constructed upon three core architectural principles:
\begin{itemize}
    \item \textbf{Ordinal over Metric Guidance:} Spatial VQA typically requires relative decisions (closer/farther) rather than absolute metrics. Ordinalization strips away unnecessary precision, enhancing robustness against pseudo-depth estimation noise.
    \item \textbf{Textual over Visual Communication:} Unlike a depth image---which enters the visual stream and is difficult for a frozen VLM to discount once encoded---a text cue shares the input channel of the question, allowing the model to naturally weigh it against the visual evidence of $I$.
    \item \textbf{Question-Targeted Filtering:} Instead of describing the entire scene geometry, DOP isolates depth evaluation strictly to the queried objects $\mathcal{O}$ and verbalizes only the specific relation required by $q$, preventing the VLM from being distracted by irrelevant background context.
\end{itemize}

\subsection{Operating Regime and Always-on Use}
\label{sec:dop-operating}
DOP targets object-level spatial questions where target categories can be localized and pseudo-depth provides reliable relative orderings. Within this regime, DOP operates in an \emph{always-on} manner---appending the ordinal cue to every eligible question without deploying any auxiliary confidence or routing models.

This setup leverages the VLM's inherent self-regulation behavior (Sec.~\ref{sec:gate}): the frozen model naturally relies on the text cue when original-image reasoning is ambiguous, yet overrides or discounts it when its internal RGB reasoning is highly confident. This always-on paradigm provides seamless assistance under visual uncertainty while remaining largely neutral when the original image already supports the answer. Explicit confidence-based gating mechanisms are evaluated in Sec.~\ref{sec:gate} but yield no performance gains over this always-on configuration.
\section{Experiments}
\label{sec:experiments}

\subsection{Experimental Setup}
\label{sec:setup}

\noindent\textbf{Models.}
Our primary backbone is \href{https://huggingface.co/Qwen/Qwen3-VL-8B-Instruct}{\textbf{Qwen3-VL-8B-Instruct}}. To test whether DOP generalizes beyond a single model, we additionally evaluate five frozen VLMs spanning two model families and two language decoders: \href{https://huggingface.co/Qwen/Qwen2-VL-7B-Instruct}{\textbf{Qwen2-VL-7B-Instruct}}, \href{https://huggingface.co/Qwen/Qwen2.5-VL-7B-Instruct}{\textbf{Qwen2.5-VL-7B-Instruct}}, \href{https://huggingface.co/llava-hf/llava-1.5-7b-hf}{\textbf{LLaVA-1.5-7B}}, \href{https://huggingface.co/llava-hf/llava-v1.6-mistral-7b-hf}{\textbf{LLaVA-1.6-Mistral-7B}}, and \href{https://huggingface.co/llava-hf/llava-v1.6-vicuna-7b-hf}{\textbf{LLaVA-1.6-Vicuna-7B}}. The LLaVA-1.6 Mistral-Vicuna pair uses the same vision stack but different LLM decoders, allowing us to test whether the effect depends on the decoder. Qwen models are run in \texttt{bfloat16} and LLaVA models in \texttt{float16}. All models are used frozen and off-the-shelf; no fine-tuning, adapter, or auxiliary module is trained.

\noindent\textbf{Depth estimators.}
For the main results, pseudo-depth is produced by \textbf{DepthAnything-v2-Large}~\cite{depthanythingv2}. To test estimator dependence, we repeat the evaluation with \textbf{DPT-Large}~\cite{dpt} and \textbf{ZoeDepth}~\cite{zoedepth}. DPT provides relative depth, while ZoeDepth provides metric depth with the opposite near/far convention; the convention is fixed a priori for each estimator. Ground-truth sensor depth from NYU Depth V2~\cite{nyudepth} and KITTI~\cite{kitti} is used only in the diagnostic quality-gradient analysis of Sec.~\ref{sec:whymapshurt}, not by DOP.

\noindent\textbf{Benchmarks.}
We evaluate on spatial-QA benchmarks that provide the queried-object localization DOP needs: \textbf{EmbSpatial-Bench}~\cite{embspatial} ($n{=}1206$; four-choice closest/farthest), \textbf{CV-Bench D+D}~\cite{cvbench} ($n{=}1200$; binary depth/distance with boxes), and a \textbf{NYU Depth V2} region-pair task~\cite{nyudepth} ($n{=}5669$ pairs, full 654-image test set; Kinect depth used only as the answer key). Object regions come from each benchmark's own annotations---EmbSpatial and CV-Bench provide the queried-object boxes, and NYU uses annotated region pairs (a markerless region task). The main results read depth at these provided boxes/regions and \emph{run no detector}; detector-supplied boxes are evaluated separately, only as an end-to-end deployment stress test (App.~\textcolor{wacvblue}{C}). The diagnostic analyses additionally use BLINK~\cite{blink}, VSR~\cite{vsr}, 3DSR-Bench~\cite{threedsrbench}, and KITTI~\cite{kitti}. We report top-1 accuracy.

\noindent\textbf{Baselines.}
We compare against seven training-free baselines: \textbf{RGB-only}; \textbf{depth-as-image} (a rendered heatmap as a second image); \textbf{depth-as-text-values} (raw per-object depths in the prompt); \textbf{chain-of-thought}; \textbf{bbox-in-prompt} (object coordinates as text); \textbf{reference-object} prompting~\cite{spatialprompt}; and \textbf{ByDeWay-LDP}~\cite{bydeway}, an always-on layered-depth captioning baseline---the closest training-free competitor. DOP is the always-on question-targeted ordinal. We also evaluate gated variants (random, zero-shot, calibrated, oracle; Sec.~\ref{sec:gate}). All methods use identical samples and decoding; prompts are in App.~\textcolor{wacvblue}{H}.

\noindent\textbf{Configuration and Cost.}
All models use greedy decoding (\texttt{do\_sample=False}) and run on a single NVIDIA H200 GPU (141\,GB). DOP needs a single VLM inference (no preceding RGB-only pass); depth maps are precomputed once per image, and likelihood reads are used only for gate ablations. The method is training-free and updates no weights, so a 7--8B backbone and the off-the-shelf depth estimator fit together in memory with ample headroom.

\subsection{Main Results}
\label{sec:results}

\begin{table*}[t]
\centering
\caption{\textbf{DOP across benchmarks and models} (generation-based; $\Delta$ vs.\ RGB-only). With the main DepthAnything-v2 estimator, DOP improves across all six backbones and three spatial benchmarks. The final row is a DPT-estimator ablation on Qwen3-VL.}
\label{tab:main}
\begin{tabular}{@{}llll@{}}
\toprule
Model & EmbSpatial (RGB$\to$DOP) & CV-Bench (RGB$\to$DOP) & NYU regions (RGB$\to$DOP) \\
\midrule
\textbf{Qwen3-VL-8B} & $69.3\to73.9$ \textbf{($+4.6$)} & $92.8\to93.7$ ($+0.9$) & $75.5\to97.2$ \textbf{($+21.7$)} \\
\textbf{Qwen2-VL-7B} & $44.0\to68.3$ \textbf{($+24.3$)} & $74.1\to75.5$ \textbf{($+1.4$)} & $56.6\to96.5$ \textbf{($+39.9$)} \\
\textbf{Qwen2.5-VL-7B} & $49.0\to70.2$ \textbf{($+21.2$)} & $78.1\to82.3$ \textbf{($+4.2$)} & $41.1\to96.6$ \textbf{($+55.5$)} \\
\textbf{LLaVA-1.6-Mistral-7B} & $46.9\to68.9$ \textbf{($+22.0$)} & $64.0\to79.2$ \textbf{($+15.2$)} & $51.8\to97.3$ \textbf{($+45.5$)} \\
\textbf{LLaVA-1.6-Vicuna-7B} & $41.9\to66.5$ \textbf{($+24.6$)} & $62.0\to80.3$ \textbf{($+18.3$)} & $49.1\to95.2$ \textbf{($+46.1$)} \\
\textbf{LLaVA-1.5-7B} & $37.5\to64.2$ \textbf{($+26.7$)} & $48.1\to54.0$ \textbf{($+5.9$)} & $44.4\to97.3$ \textbf{($+52.9$)} \\
\textbf{Qwen3} / \textbf{DPT estimator} & $69.3\to73.2$ \textbf{($+3.9$)} & $92.8\to92.5$ ($-0.3$) & $75.5\to90.2$ \textbf{($+14.7$)} \\
\bottomrule
\end{tabular}
\end{table*}

\noindent\textbf{DOP improves when depth is informative and remains neutral when the original image is already strong.}
Table~\ref{tab:main} reports DOP across all six backbones and three benchmarks (DepthAnything-v2). On our primary \textbf{Qwen3-VL-8B}, DOP adds $+4.6$pp on EmbSpatial ($69.3\to73.9$) and $+21.7$pp on the NYU region task ($75.5\to97.2$), while staying neutral on CV-Bench ($92.8\to93.7$, $+0.9$pp) where the original image already suffices. The gains are largest exactly where original-image reasoning is weak---$+24.3$pp (Qwen2-VL) and $+22.0$pp (LLaVA-1.6-Mistral) on EmbSpatial, and over $+45$pp on NYU for several weaker backbones (Table~\ref{tab:main})---and the same trend holds under a different estimator (DPT: $+3.9$pp on EmbSpatial, $-0.3$pp on CV-Bench; last row of Table~\ref{tab:main}).

\noindent\textbf{DOP is competitive with the strongest training-free depth-prompting baseline.}
On EmbSpatial (Table~\ref{tab:baselines}), DOP reaches $73.9\%$---$+4.6$pp over RGB-only ($69.3\%$)---and significantly beats five of six training-free baselines under paired McNemar with Holm correction (all $p<0.001$). It \emph{ties} ByDeWay-LDP ($73.0\%$; $+0.9$pp, $p{=}0.15$, n.s.), the closest competitor, with a far simpler cue: one queried-object ordinal instead of ByDeWay's verbose global captions. On CV-Bench (Table~\ref{tab:cvbench}), a strong-RGB regime ($92.8\%$), DOP is slightly positive ($+0.9$pp) while raw depth values ($-8.8$pp) and the depth image ($-0.7$pp) hurt---the small positive margins should be read cautiously, but the large drops confirm our interface diagnosis. Across the full 15-cell cross-model sweep (App.~\textcolor{wacvblue}{I}), the two best conditions are always the depth-\emph{text} cues (DOP, ByDeWay), never an image or raw numbers.

\noindent\textbf{Boundary conditions.}
DOP's operating regime is object-level questions with localizable targets. The one clear failure is BLINK Relative Depth, whose abstract point markers push pseudo-depth toward chance so the forced ordinal misleads ($-22.6$pp; ordering accuracy $58\%$, near chance, App.~\textcolor{wacvblue}{R}). On two \emph{box-free} depth benchmarks (VSR, 3DSR-Bench), where DOP can only emit a degraded scene-level cue rather than its targeted per-object ordinal, it stays largely neutral with the strong backbone (VSR $+0.6$, 3DSR-Bench $+0.4$; App.~\textcolor{wacvblue}{Q})---the missing boxes explain the small gains, and it does not materially hurt.

\begin{table}[t]
\centering
\caption{\textbf{Baselines and gate ablations} (EmbSpatial, Qwen3, $n=1206$). 95\% CIs are bootstrap intervals; significance is paired McNemar vs.\ DOP with Holm correction (\textbf{***}: $p<0.001$; n.s.: not significant).}
\label{tab:baselines}
\resizebox{\linewidth}{!}{%
\begin{tabular}{@{}llll@{}}
\toprule
Condition & Acc.\ [95\% CI] & $\Delta$ RGB & DOP vs.\ it \\
\midrule
rgb\_only & $69.3,[66.7, 71.8]$ & --- & $+4.6$, $p{=}5.9{\times}10^{-4}$ *** \\
depth-as-image (RGB+heatmap) & $69.4,[66.8, 72.0]$ & $+0.1$ & $+4.5$, $p{=}7.0{\times}10^{-4}$ *** \\
depth-as-text-values & $53.8,[51.2, 56.7]$ & $\mathbf{-15.5}$ & $+20.1$, $p{<}10^{-37}$ *** \\
CoT (fair, 512 tok) & $66.3$ & $-3.0$ & --- \\
bbox-in-prompt & $66.5,[63.8, 69.1]$ & $-2.8$ & $+7.4$, $p{<}10^{-6}$ *** \\
reference-object & $67.5,[64.8, 70.1]$ & $-1.8$ & $+6.4$, $p{<}10^{-5}$ *** \\
ByDeWay-LDP & $73.0,[70.4, 75.5]$ & $+3.6$ & $+0.9$, $p{=}0.15$ n.s. \\
\textbf{DOP (ours)} & $\mathbf{73.9,[71.1, 76.0]}$ & $\mathbf{+4.6}$ & --- \\
\midrule
\multicolumn{4}{@{}l}{\emph{Gate ablations (Sec.~\ref{sec:gate}); none beats always-on DOP.}} \\
two-sided gate (depth vs. model margin) & $72.6$ & $+3.5$ & $<$ DOP \\
random 50\% gate & $72.1$ & $+3.0$ & $<$ DOP \\
oracle gate (upper bound) & $83.7$ & $+14.6$ & upper bound \\
\bottomrule
\end{tabular}}
\end{table}

\noindent\textbf{Comparison to trained depth-specialist VLMs.}
Trained depth-aware VLMs use model-specific data and formats, so they are complementary rather than plug-and-play baselines: SpatialBot-3B~\cite{spatialbot} expects its own RGB+depth format and, under forced-choice questions, emits free-form text, so its accuracy reflects format mismatch, not depth ability (App.~\textcolor{wacvblue}{M}). DOP instead asks how far a \emph{frozen} general VLM goes with a one-line cue and no training.

\begin{table}[t]
\centering
\caption{\textbf{Baselines on CV-Bench} (Qwen3, $n=1200$). This is a strong original-image regime: RGB-only already reaches $92.8\%$. DOP remains largely neutral and slightly positive, while raw depth values hurt.}
\label{tab:cvbench}
\resizebox{\linewidth}{!}{%
\begin{tabular}{@{}lll@{}}
\toprule
Condition & Acc. (\%) & $\Delta$ vs. RGB \\
\midrule
rgb\_only & $92.8$ & --- \\
depth-as-image (RGB+heatmap) & $92.1$ & $-0.7$ \\
depth-as-text-values & $83.9$ & $\mathbf{-8.8}$ \\
CoT prompting & $92.1$ & $-0.7$ \\
bbox-in-prompt & $91.6$ & $-1.2$ \\
reference-object & $92.2$ & $-0.6$ \\
ByDeWay-LDP & $93.5$ & $+0.8$ \\
\textbf{DOP (ours)} & $\mathbf{93.7}$ & $\mathbf{+0.9}$ \\
\bottomrule
\end{tabular}}
\end{table}

\begin{figure*}[t]
\centering
\begin{subfigure}[t]{0.245\linewidth}
\includegraphics[width=\linewidth]{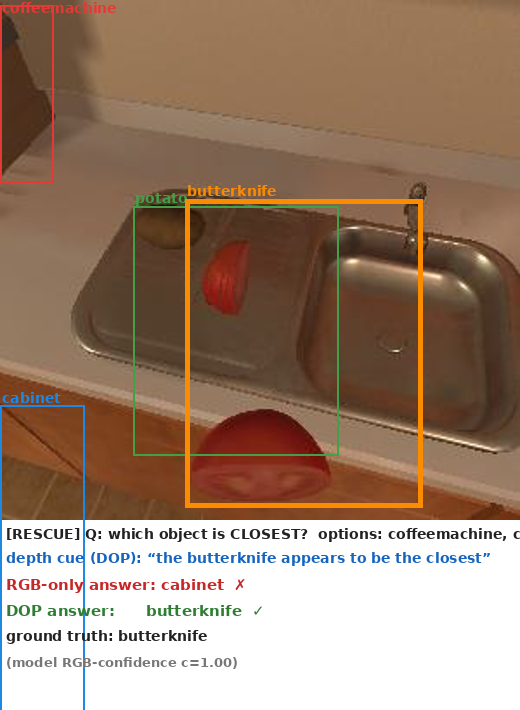}
\caption{Rescue: RGB ``cabinet'' $\to$ DOP \textbf{``butterknife''} (correct).}
\end{subfigure}\hfill
\begin{subfigure}[t]{0.245\linewidth}
\includegraphics[width=\linewidth]{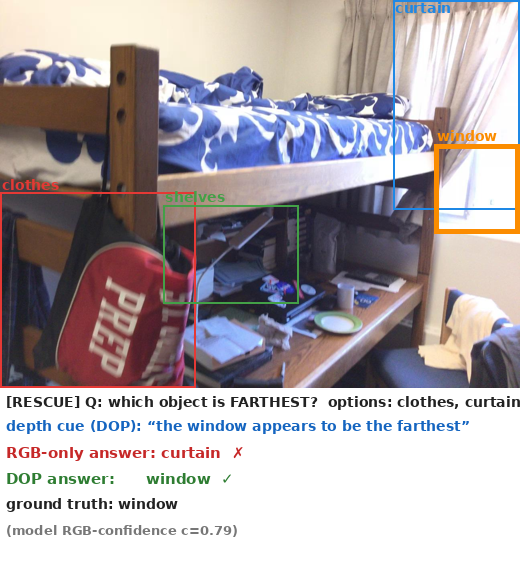}
\caption{Rescue: RGB ``curtain'' $\to$ DOP \textbf{``window''} (correct).}
\end{subfigure}\hfill
\begin{subfigure}[t]{0.245\linewidth}
\includegraphics[width=\linewidth]{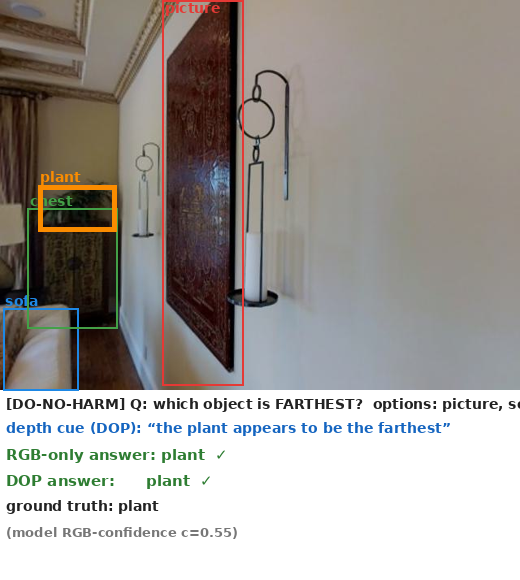}
\caption{Neutral: RGB \textbf{``plant''} stays correct under DOP.}
\end{subfigure}\hfill
\begin{subfigure}[t]{0.245\linewidth}
\includegraphics[width=\linewidth]{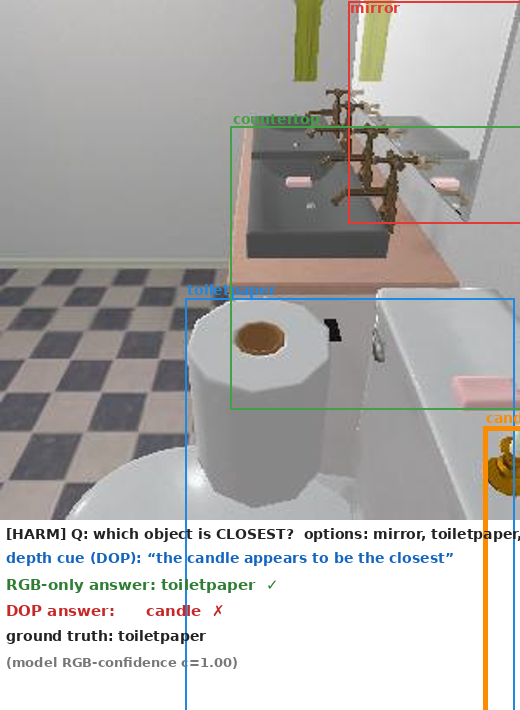}
\caption{Harm case: RGB \textbf{``toiletpaper''} $\to$ DOP ``candle'' (wrong).}
\end{subfigure}

\caption{\textbf{Qualitative behavior of DOP} (EmbSpatial, Qwen3-VL). DOP can rescue original-image errors on queried objects (a,b), preserve already-correct answers (c), and occasionally harm when the depth cue is misleading (d). Overall, DOP rescues RGB errors $1.5\times$ as often as it harms correct answers (Sec.~\ref{sec:gate}).}
\label{fig:qualitative}
\end{figure*}

\subsection{Self-Regulation and Gate Ablations}
\label{sec:gate}
\label{sec:selfreg}

DOP is intentionally always-on: should the cue instead be gated by confidence? No (Table~\ref{tab:baselines}, bottom). Always-on DOP ($73.9\%$) beats every gate we tried---a two-sided depth-vs-answer-margin gate ($72.6\%$), a random $50\%$ gate ($72.1\%$), and single-signal depth-only ($71.9\%$) or model-only ($71.0\%$) gates---while an oracle gate reaches $83.7\%$, so real headroom exists but no zero-shot reliability signal we tried captures it.

The reason is that the frozen VLM already self-regulates the text cue rather than copying it. The clearest evidence is a \emph{wrong-cue} test: on the $17.6\%$ of CV-Bench items where the pseudo-depth ordinal is actually \emph{wrong}, DOP still scores $84.4\%$---a model that blindly followed the cue would score $0\%$ there---losing only $5.7$pp versus RGB-only on those items. Consistently, cue adoption is higher when the model is uncertain and falls as its RGB-answer confidence rises ($46\%\to27\%$, $r=-0.12$, $p<10^{-4}$); harm concentrates on low-confidence items, and DOP rescues RGB errors $1.5\times$ as often as it harms correct ones. A causal attention-knockout (App.~\textcolor{wacvblue}{A}) leaves high-confidence answers nearly unchanged ($-0.6$pp) while strongly perturbing unsure ones---the model leans on the cue precisely when its own reading is weak, not blindly.

\begin{table}[t]
\centering
\caption{\textbf{Robustness} (EmbSpatial, Qwen3, $n=600$). DOP is robust to moderate bounding-box jitter and to cue wording.}
\label{tab:robust}
\resizebox{\linewidth}{!}{%
\begin{tabular}{@{}lrrrr@{}}
\toprule
bbox jitter (frac) & $0.0$ & $0.1$ & $0.2$ & $0.3$ \\
DOP $\Delta$ vs. RGB & $+4.5$ & $+4.2$ & $+2.2$ & $-1.8$ \\
\midrule
cue wording & plain & camera & ranking & estimator \\
DOP $\Delta$ vs. RGB & $+4.5$ & $+4.0$ & $+5.8$ & $+3.2$ \\
\bottomrule
\end{tabular}}
\end{table}

\subsection{Robustness}
\label{sec:robust}

DOP needs neither perfect localization nor a hand-tuned phrase (Table~\ref{tab:robust}): it stays net-positive under bounding-box jitter up to $\pm20\%$ and across four cue phrasings (plain, camera-centric, ranking, estimator-centric). It turns negative ($-1.8$pp) only under severe $\pm30\%$ jitter, where the pooling window leaks into background or adjacent objects and corrupts the depth readout---so DOP is robust to moderate localization noise and is not tied to a single cue wording. End-to-end detector quality is a separate deployment bottleneck (appendix).

\begin{table}[t]
\centering
\caption{\textbf{Irrelevant-depth stress test} (Qwen3-VL; $\Delta$ vs.\ RGB-only). On four benign benchmarks (POPE/AMBER/MME/CHAIR) the irrelevant \emph{text} cue is largely harmless while the \emph{same} depth as an \emph{image} disrupts every one; HallusionBench is the adversarial exception where every cue hurts. Form, not content, drives the harm.}
\label{tab:donoharm}
\resizebox{\linewidth}{!}{%
\begin{tabular}{@{}llrrr@{}}
\toprule
Benchmark & Metric & RGB & DOP ($\Delta$) & depth-img ($\Delta$) \\
\midrule
POPE & acc & $89.1$ & $89.7$ \textbf{($+0.6$)} & $88.5$ ($-0.7$) \\
AMBER & acc & $89.4$ & $88.9$ \textbf{($-0.6$)} & $85.1$ ($-4.3$) \\
MME & score & $2392.7$ & $2422.4$ \textbf{($+29.7$)} & $2365.2$ ($-27.5$) \\
CHAIR & CHAIR\_s$\downarrow$ & $11.2$ & $\mathbf{6.6}$ \textbf{($-4.6$)} & $40.0$ ($\mathbf{+28.8}$) \\
\midrule
HallusionBench & aAcc & $81.7$ & $75.2$ ($-6.6$) & $76.3$ ($-5.4$) \\
\bottomrule
\end{tabular}}
\end{table}

\subsection{Irrelevant-Cue Stress Test Beyond Spatial Reasoning}
\label{sec:donoharm-general}

An always-on cue should not corrupt core abilities when depth is irrelevant. We add the \emph{same} depth---as a DOP text cue or as a second image---to four general benchmarks: POPE~\cite{pope}, AMBER~\cite{amber}, MME~\cite{mme}, and CHAIR~\cite{chair} (Table~\ref{tab:donoharm}). The text cue stays benign and even helps (MME $+29.7$, $\text{CHAIR}_s$ $-4.6$), whereas the depth \emph{image} disrupts every task (AMBER $-4.3$, $\text{CHAIR}_s$ $+28.8$). The same content, opposite sign by form---direct evidence that the bottleneck is the interface channel, not depth content.

\noindent\textbf{Adversarial Exception.}
HallusionBench~\cite{hallusionbench} marks the boundary: built from visual illusions and false-premise traps, it degrades \emph{every} cue, DOP most ($-6.6$pp; Table~\ref{tab:donoharm}). Self-regulation in benign settings does not imply safety against adversarial premises---a confident linguistic cue can worsen fragile predictions when the question itself is engineered to mislead.

\section{Conclusion}
\label{sec:conclusion}

We show that depth is not merely missing from frozen VLMs; rather, its usefulness depends on the interface through which it is provided. Showing monocular depth as an additional image can burden the model with visual alignment, depth-map interpretation, and noisy fine-grained ordering. We therefore propose Depth-Ordinal Prompting (DOP), a training-free method that converts depth into a single question-targeted ordinal text cue at the queried objects. Across models, benchmarks, and depth estimators, DOP improves spatial reasoning when pseudo-depth provides reliable object-level ordering and remains largely neutral when the original image is already sufficient. We provide further details in the appendix, including self-regulation analyses, detector-based localization stress tests, BLINK and HallusionBench boundary cases, full cross-model sweeps, prompt templates, and qualitative examples, to make the scope and limitations of DOP explicit.


{
    \small
    \bibliographystyle{ieeenat_fullname}
    \bibliography{main}
}

\end{document}